\pdfoutput=1
\documentclass[11pt]{article}
\usepackage{fullpage,times}
\usepackage{parskip}
\usepackage{titling}

\usepackage{amsmath, amsthm, amssymb, amsfonts, mathtools, graphicx, enumerate}
\usepackage{adjustbox}
\usepackage[utf8]{inputenc} 
\usepackage[T1]{fontenc}    
\usepackage{url}            
\usepackage{booktabs}
\usepackage{nicefrac}       
\usepackage{microtype}      
\usepackage{xspace}
\usepackage{algorithm,algorithmic}
\usepackage{color}
\usepackage{enumitem}
\usepackage{comment}
\usepackage{bm}
\usepackage{apptools}
\usepackage[page, header]{appendix}
\usepackage{titletoc}
\usepackage{array}

\usepackage[scaled=.9]{helvet}

\usepackage{url}
\usepackage{caption}
\usepackage[table,dvipsnames]{xcolor}
\usepackage[colorlinks=true, linkcolor=blue, citecolor=blue]{hyperref}

\usepackage{natbib}
\bibpunct{(}{)}{;}{a}{,}{,}

\usepackage{amsthm}

\begingroup
    \makeatletter
    \@for\theoremstyle:=definition,remark,plain\do{%
        \expandafter\g@addto@macro\csname th@\theoremstyle\endcsname{%
            \addtolength\thm@preskip\parskip
            }%
        }
\endgroup

\newcommand{\myempty}[1]{}

\newcommand{\yrm}[1]{}

\newcommand{\qiangremoved}[1]{}

 \def\bb#1\ee{\begin{align*}#1\end{align*}}

 \def\bba#1\eea{\begin{align}#1\end{align}}

\pretitle{\centering\LARGE}
\posttitle{\par\vskip 0.3em}
\predate{\centering\large}
\postdate{\par\vskip -1em}
\hypersetup{
  pdftitle={Admissible Diffusion for Multimodal Interventional Trajectories},
  pdfauthor={Xing Han, Shravan Chaudhari, Jiarui Shao, Paul Pu Liang, Suchi Saria},
  pdfsubject={A framework and preliminary experiments for clinical trajectory generation},
  pdfkeywords={causal inference, multimodal learning, diffusion, clinical trajectories}
}

\title{\huge Admissible Diffusion for \\ Multimodal Interventional Trajectories}
\author{%
  Xing Han\textsuperscript{1,*}\quad
  Shravan Chaudhari\textsuperscript{1,*}\quad
  Jiarui Shao\textsuperscript{1}\\[3pt]
  Paul Pu Liang\textsuperscript{2}\quad
  Suchi Saria\textsuperscript{1,3}\\[5pt]
  \small\textsuperscript{1}Johns Hopkins University\quad
  \small\textsuperscript{2}Massachusetts Institute of Technology\\
  \small\textsuperscript{3}Bayesian Health
}
\date{}

\begin{document}
\maketitle
\begingroup
\renewcommand{\thefootnote}{\fnsymbol{footnote}}
\footnotetext[1]{Xing Han and Shravan Chaudhari contributed equally to this work.}
\endgroup

\begin{abstract}
Generating a plausible clinical trajectory does not establish what would happen under a different treatment. We present ADMIT, a framework combining irregular multimodal representations, treatment-conditioned latent diffusion and explicit constraints on generated states or actions. We formulate its interventional target through sequential g-computation and distinguish causal assumptions from constraint satisfaction. Its admissibility mechanism translates physiological prior knowledge into explicit constraints on generated states and proposed actions. Treatment-exposure dynamics condition latent transitions, while state projection or action gating applies the constraints during rollout so that they influence subsequent trajectory generation. In our preliminary experiments, multimodal inputs improved supervised hidden-state recovery and reduced treatment-contrast error. In a simulated dosing-schedule experiment with leak-free history encoding, ADMIT predicted most of the tumor-volume change caused by redistributing a fixed total dose. An exposure input improved these predictions around a temporary dose reduction whether or not the assumed clearance rate was correct, but reduced the predicted size of a dose effect, and a deterministic recurrent baseline matched ADMIT's average predictions. Exposure projection reduced constraint violations, although enforcement remained incomplete. Semi-synthetic experiments using eICU context illustrated treatment-response generation under fixed and adaptive policies. Observational examples further characterize model treatment sensitivity. ADMIT provides a framework for testing whether complementary observations and physiological restrictions improve intervention trajectories, with representation recovery, effect accuracy and rule enforcement assessed separately.

\end{abstract}

\section{Introduction}
Treatment changes a patient's subsequent condition, which can in turn determine the next treatment. A trajectory model used to compare clinical strategies must account for this feedback: predictions under the observed treatment process need not describe outcomes under a different plan \citep{robins1986causal,schulam2017reliable}. The object of interest is therefore a distribution of future states under a specified sequence of decisions, conditional on the history available when the comparison begins.
That history can contain irregular measurements from several modalities. Compressing these observations into a latent state introduces another requirement: information needed to adjust for treatment assignment must survive compression. Representation-induced confounding is a documented concern in treatment-effect estimation \citep{melnychuk2024ricb}. A complementary problem arises after generation. A model may produce trajectories that violate a known exposure recurrence or an investigator-specified bound. Physical restrictions can constrain a diffusion model \citep{bastek2025physics}, but a plausible trajectory still needs causal assumptions and adequate treatment support before it can answer an intervention question \citep{robins1986causal}.

Multimodal fusion and foundation models support clinical prediction, question answering and report generation \citep{han2024fusemoe,zhang2024biomedgpt}; treatment-dependent trajectory simulation is not among their reported evaluations. Longitudinal causal models can generate counterfactual trajectories, but their evaluations focus on relatively low-dimensional structured variables: CRN and Causal Transformer predict selected outcome sequences \citep{bica2020crn,melnychuk2022causal}, while G-Transformer also simulates multivariate covariate trajectories \citep{xiong2024gtransformer}. These demonstrations do not include generation of images or clinical narratives. Medical event generators such as Curiosity and EHRWorld generate sequences of clinical events \citep{waxler2025curiosity,mu2026ehrworld}, but their reported generation procedures do not impose explicit physiological constraints on actions and state transitions. These gaps call for a simulator that addresses all three challenges together: learning from multimodal histories, generating trajectories under alternative treatment plans and applying physiological knowledge during rollout.

\begin{figure}[t]
 \centering
 \includegraphics[width=0.80\linewidth]{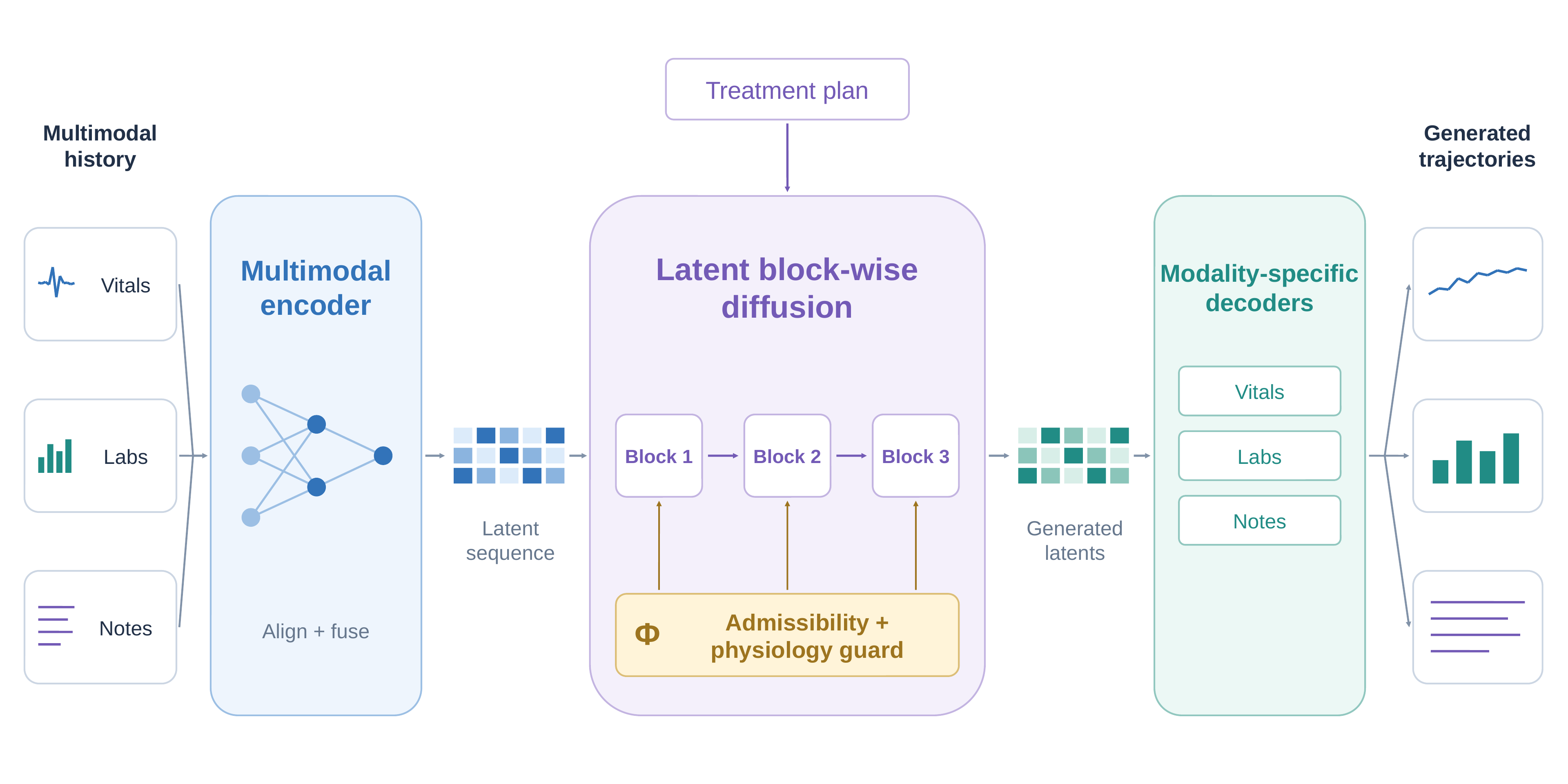}
 \caption{\textbf{ADMIT framework.} Aligned multimodal observations form latent states. Treatment-conditioned diffusion generates successive blocks, with optional constraints before updating the history. Modality-specific heads reconstruct inputs and decode selected targets. The experiments generate numerical outputs; the notes channel uses categorical tokens. The physiology guard approximately enforces a specified rule.}
 \label{fig:overview}
\end{figure}

We present ADMIT (Admissible Diffusion for Multimodal Interventional Trajectories), a framework for simulating how a patient's condition may evolve under alternative treatment plans using multimodal histories and specified physiological constraints (Fig.~\ref{fig:overview}). It builds on sequential g-computation and conditional diffusion \citep{xiong2024gtransformer,alinezhad2026cdm}, with constraints applied during rollout. Its central contribution is to make physiological admissibility part of intervention simulation: prior knowledge constrains proposed actions and generated state transitions as a trajectory unfolds. The intended clinical use is to compare possible consequences of alternative care plans for the same patient and inspect their consistency with specified physiological rules.

We ask whether complementary observations and mechanistic restrictions improve generated intervention trajectories. Synthetic experiments test representation recovery, predictions under changed dosing schedules with absent, correct or misspecified drug-clearance input, and constraint enforcement; real eICU histories paired with known treatment-response equations provide a semi-synthetic test of counterfactual prediction and treatment-effect estimation; observational clinical examples measure model sensitivity where counterfactual truth is unavailable. Together, these evaluations are designed to test whether physiological priors improve counterfactual accuracy while reducing violations of specified constraints.

\section{Related work}
Longitudinal counterfactual prediction includes recurrent models with adversarially balanced representations \citep{bica2020crn}, attention-based treatment and outcome representations \citep{melnychuk2022causal}, and g-computation models that simulate multivariate trajectories under dynamic regimens \citep{xiong2024gtransformer}. G-Transformer is particularly relevant because its Monte Carlo rollout explicitly updates evolving covariates under a queried policy, a capability shared with the intended ADMIT specification \citep{xiong2024gtransformer}. Balanced representations must still preserve information needed for adjustment \citep{melnychuk2024ricb}.

Diffusion models provide flexible conditional distributions. CSDI applies score-based generation to probabilistic time-series imputation \citep{tashiro2021csdi}; causal diffusion models explicitly target longitudinal counterfactual outcome distributions \citep{alinezhad2026cdm}. The latter report simulated one-step tumor-outcome evaluations conditioned on history. Our preliminary evaluation emphasizes longer rollouts and explicit constraint operations; this difference in evaluation scope does not imply a restriction on their generative architecture \citep{alinezhad2026cdm}. ADMIT uses diffusion within a treatment-conditioned trajectory model and exposes mechanistic restrictions during rollout. Such restrictions have precedents in physics-informed diffusion \citep{bastek2025physics}. Its multimodal representation draws on time attention for irregular observations \citep{shukla2021mtan} and mixture-of-experts fusion \citep{han2024fusemoe}. The proposed organization allows the intervention plan, learned transition and imposed constraint to be inspected separately, including when the constraint changes the meaning of the generated trajectory.

Multimodal foundation models for healthcare seek reusable representations of heterogeneous clinical data. BiomedGPT \citep{zhang2024biomedgpt} and Med-PaLM M \citep{tu2024medpalmm} support tasks such as medical image interpretation, question answering and report generation. MERGE models interactions across modalities \citep{han2026merge}, and FLAME supports continual adaptation across clinical tasks and modality combinations \citep{han2026flame}. ADMIT focuses on using multimodal patient states to generate trajectories under alternative treatment plans while applying explicit physiological constraints.

\section{Method}
\label{sec:method}
Let $L_t$ denote the measured state before treatment $A_t$, and let $H_t=(L_{0:t},A_{0:t-1})$ be the available history. A query specifies either a fixed future plan or a policy $\pi(a_t\mid H_t)$ evaluated on generated history. Under consistency, sequential exchangeability and positivity, the longitudinal g-formula identifies the joint interventional law \citep{robins1986causal}:
\begin{equation}
 p^{\pi}(l_{s+1:T},a_{s:T-1}\mid H_s)
 =\prod_{t=s}^{T-1}p(l_{t+1}\mid H_t,a_t)\,\pi(a_t\mid H_t).
 \label{eq:gformula}
\end{equation}
Marginalizing actions gives the state distribution of interest. Replacing the measured history with a learned representation additionally requires that the representation retain adequate adjustment and predictive information and that its transition law be estimated accurately. These conditions are assumptions, not consequences of multimodal input or diffusion. The target is a history-conditional interventional distribution; recovering a particular patient's unobserved counterfactual noise requires additional assumptions about how potential outcomes are coupled.

Each modality supplies values, times and observation masks. Linear or lookup embeddings represent continuous or categorical values. Time attention aligns observations to a reference grid \citep{shukla2021mtan}; a gated local kernel retains rapid variation. A top-two mixture of four experts fuses these representations \citep{han2024fusemoe}, and Gaussian variational heads produce a 16-dimensional latent state. Stage~1 minimizes masked Gaussian or categorical reconstruction losses, a Kullback--Leibler term and an expert-balancing term. Clinical masks indicate retained rows after imputation, rather than original per-variable availability. Optional exposure and comorbidity heads require simulator labels. The encoder and decoders are then frozen, and standardized posterior means provide transition-training targets.

For the causal specification, $Z_t$ must use only observations available by $t$, with $R_t=\psi(Z_{0:t},A_{0:t-1})$ summarizing pre-action history. The evaluated encoder instead attends to the complete factual record before extracting a prefix; future observations can therefore enter its seed. The dosing-schedule experiment (Section~\ref{sec:p1}) instead marks observations after the history as missing and sets them to zero before encoding, so its history uses only past observations. The implemented GRU summary also includes the current action. We report the other runs as conditional-generation diagnostics, not prospective forecasting evidence.

The transition model generates a block of residuals $U_0=Z_{t+1:t+B}-Z_t$, anchored at the final history state. Let $Q_t$ contain the history summary, anchor, proposed actions and optional exposure features. Following DDPM training \citep{ho2020ddpm},
\begin{equation}
 U_k=\sqrt{\bar\alpha_k}U_0+\sqrt{1-\bar\alpha_k}\,\epsilon,
 \qquad
 \mathcal L_{\rm diff}=\mathbb E\!\left[\|\epsilon-\epsilon_\theta(U_k,k;Q_t)\|_2^2\right],
 \label{eq:diffusion}
\end{equation}
where $\epsilon\sim\mathcal N(0,I)$ and $\bar\alpha_k$ is the cumulative noise-schedule coefficient. Actions modulate denoiser features through learned scale and shift parameters. An optional free-running loss penalizes factual latent errors after feeding generated states back into history. Outcomes are decoded from sampled trajectories. Their spread requires separate calibration; disabling reverse-step noise still leaves random initialization and does not yield a conditional mean.

Rollout samples future patient trajectories under a specified treatment plan by repeatedly applying the learned transition to the evolving history. Actions come from a fixed plan or an adaptive policy evaluated on generated states. Mechanistic knowledge can enter through additional state variables, such as drug exposure, that summarize treatment carryover and condition future transitions. Optional admissibility checks constrain proposed actions or generated states before they influence subsequent predictions. Appendix~\ref{app:rollout-exposure} gives the rollout procedure and exposure specification.

To implement Eq.~\eqref{eq:gformula} for adaptive treatment, states and actions must alternate at the decision times. A one-step transition ($B=1$), or an explicitly sequential factorization within each block, supports this ordering. A joint block conditioned on every realized action requires stronger assignment assumptions. For two steps,
\begin{equation}
 p(l_1,l_2\mid h_0,a_0,a_1)\propto
 p(l_1\mid h_0,a_0)\,p(a_1\mid h_0,a_0,l_1)\,p(l_2\mid h_0,a_0,l_1,a_1).
 \label{eq:blockselection}
\end{equation}
The middle factor retains the observational assignment mechanism when a new action block is substituted. Joint-block identification instead requires plan assignment from block-start history with appropriate exchangeability and support.

An admissibility rule is a declared constraint $g(D(z),a,H)\leq0$ on decoded states or actions. The implemented concentration projector iterates
\begin{equation}
 z\leftarrow z-\frac{[\widehat C(z)-C_{\max}]_+}
 {\|\nabla_z\widehat C(z)\|_2^2+\varepsilon}\nabla_z\widehat C(z),
 \label{eq:projection}
\end{equation}
and appends the transformed state. Finite iterations can leave violations, requiring a residual check. Projection changes the transition law and therefore the original intervention distribution. Alternatively, an action gate suppresses doses exceeding an exposure threshold and defines a modified policy. Deterministic exposure must remain consistent with the administered actions. Broader physiological training penalties remain an extension. A toxicity threshold is an investigator-defined limit, not physical impossibility; satisfying it neither establishes causal identification nor certifies clinical safety.

\section{Experiments}
\label{sec:experiments}
We evaluate ADMIT across synthetic, semi-synthetic and observational settings. Using a tumor simulator adapted from \citet{geng2017tumor}, we test whether complementary modalities improve recovery of a hidden confounder $H$ and prediction of treatment contrasts, whether exposure priors improve dose-timing predictions, and whether an exposure limit reduces rule violations. Semi-synthetic experiments test treatment-response generation in real clinical context. Observational examples assess sensitivity to treatment queries without known counterfactual outcomes.

\begin{table}[t]
\centering
\small
\caption{\textbf{Synthetic representation and treatment-contrast diagnostics.} Mean $\pm$ sample standard deviation across three seeds. Visibility zero removes the direct contribution of $H$ to target observations. The $H$ readout is jointly supervised. Treatment contrasts compare always treating with never treating, pooled over 20 future steps and three standardized target channels.}
\label{tab:synthetic}
\begin{tabular}{llccc}
\toprule
$H$ visibility & Inputs & $H$ MSE & Contrast RMSE & Contrast correlation \\
\midrule
1 & Multimodal & $0.130\pm0.007$ & $0.755\pm0.022$ & $0.838\pm0.010$ \\
1 & Target only & $0.159\pm0.016$ & $0.783\pm0.014$ & $0.838\pm0.004$ \\
0 & Multimodal & $0.203\pm0.012$ & $0.783\pm0.008$ & $0.843\pm0.016$ \\
0 & Target only & $0.784\pm0.045$ & $0.913\pm0.034$ & $0.827\pm0.004$ \\
\bottomrule
\end{tabular}
\end{table}

\begin{figure}[t]
\centering
\includegraphics[width=\linewidth]{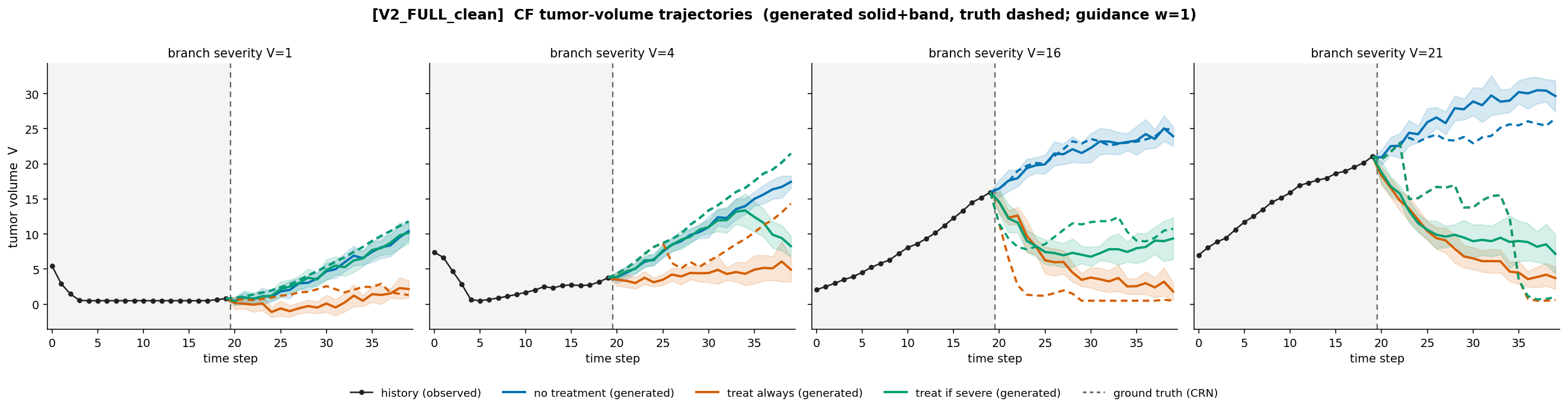}
\caption{\textbf{Synthetic tumor trajectories.} Four examples span severity at intervention after 20 steps. Solid lines and bands show generated means and sample standard deviations; dashed lines show paired simulator trajectories. Treatment plans never treat, always treat, or treat when tumor volume exceeds 12. The model updates the adaptive plan once per generated block; the simulator updates it every step. Black history points denote simulator states. The model encodes the full observed record, including future observations, so these examples do not establish forecasting accuracy.}
\label{fig:trajectories}
\end{figure}

When $H$ is hidden from the target, multimodal inputs reduce its supervised recovery MSE by 74.2\% and contrast RMSE by 14.2\% (Table~\ref{tab:synthetic}), although this improvement does not verify that the representation preserves every required confounder. Figure~\ref{fig:trajectories} illustrates policy-specific trajectories against paired simulator references sharing process noise. The full-record encoding limitation applies to both evaluations.

\subsection{Predicting the effects of dose timing and drug clearance}
\label{sec:p1}
The same total dose can be delivered steadily, concentrated early, or reduced temporarily. We examine whether ADMIT can predict how these dosing choices change tumor growth. We use the tumor simulator to isolate this question, then examine whether knowledge of drug clearance improves the predictions and whether a simpler model performs as well.

This experiment uses one drug, with doses between 0 and 1, and switches off the hidden comorbidity. Treatment follows courses that do not depend on tumor volume, allowing us to study dose timing without confounding. Each patient has 12 observed steps followed by 12 forecast steps. All conditions use the same 1,000 training, 200 validation and 300 test patients, with three training seeds that vary model initialization and batch order. Only observations available before forecasting enter the encoded history: later observations are marked as missing and set to zero. Replacing them with another patient's measurements leaves that history unchanged (Appendix~\ref{app:p1}).

\begin{figure}[!t]
\centering
\includegraphics[width=\linewidth]{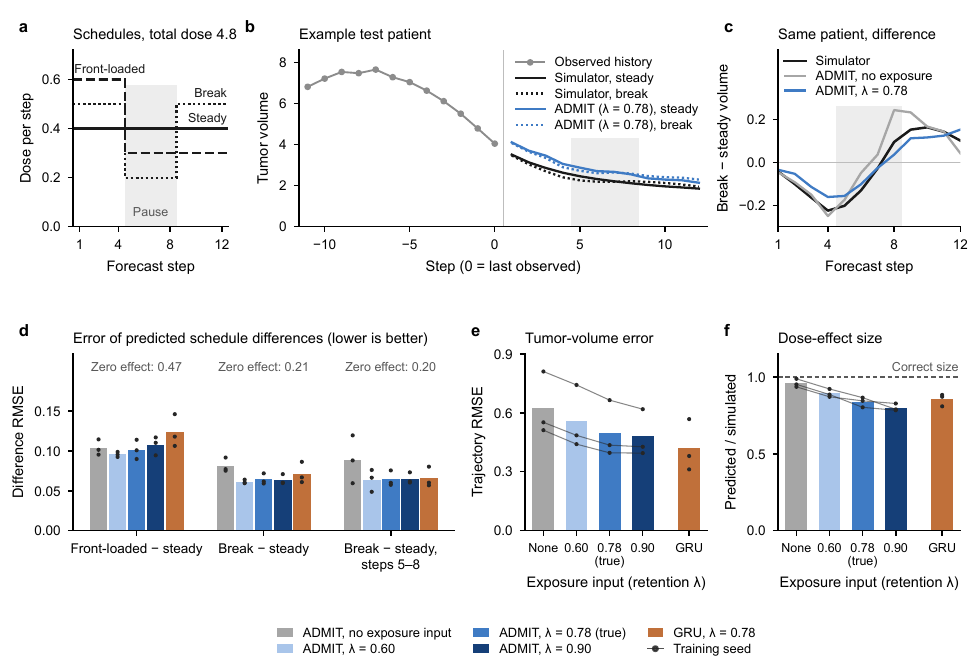}
\caption{\textbf{Dosing schedules and drug clearance in the tumor simulator.} (a) Forecast schedules, each with total dose 4.8; shading marks the reduced-dose period of the break schedule (steps 5--8) in panels a--c, labeled ``Pause in panel a. (b) One test patient, chosen before viewing predictions as the patient whose simulated break-minus-steady difference is of median size: observed history, simulator references and ADMIT with the true fraction of drug retained per step ($\lambda=0.78$, training seed 0) under steady and break dosing. (c) The same patient's break-minus-steady difference in tumor volume. (d) RMSE of predicted schedule differences; text gives the zero-effect reference. (e) RMSE of predicted tumor volume, averaged over the three schedules. (f) Predicted divided by simulated difference between sustained doses of 0.6 and 0.2; the dashed line marks the correct size. In d--f, bars show means over three training seeds, dots show individual seeds, and lines in e and f connect the same seed across retention values; each seed scores 300 test patients.}
\label{fig:p1}
\end{figure}

We compare three schedules, each giving a total dose of 4.8 (Fig.~\ref{fig:p1}a). \emph{Steady} dosing gives 0.4 at every step; \emph{front-loaded} dosing gives 0.6 for four steps and 0.3 thereafter; and the \emph{break} schedule gives 0.5, reduces the dose to 0.2 for steps 5--8, then returns to 0.5. Thus, the break is a temporary dose reduction. The dose levels and changes are represented in training (Appendix~\ref{app:p1}). For each patient, we measure how tumor volume under each alternative schedule differs from steady dosing. We score the error in these differences against the simulator and compare it with a \emph{zero-effect reference}, which predicts that changing the schedule makes no difference. Separately, we score error in tumor volume itself and test the predicted benefit of increasing sustained dosing from 0.2 to 0.6.

\emph{Drug exposure} is the amount of drug remaining in the body after current and earlier doses. The simulator retains about 78\% of exposure from one step to the next ($\lambda=0.78$). We give ADMIT either no exposure input, exposure calculated with this correct retention, or exposure calculated with faster ($\lambda=0.60$) or slower ($\lambda=0.90$) clearance. All other ADMIT settings are shared. We also compare a gated recurrent unit (GRU) network, a simpler model that uses the same patient representation, doses and correct exposure input to produce one predicted course. ADMIT's predictions average 64 generated courses; simulator references average 32 runs, with random variation shared across schedules within each method. Training settings and metric definitions are given in Appendix~\ref{app:p1}.

\begin{table}[t]
\centering
\small
\caption{\textbf{Predicting the effects of dose timing and dose size.} Mean $\pm$ sample standard deviation across three training seeds, each scoring the same 300 test patients. The first three columns give the RMSE of predicted schedule differences and the fourth the RMSE of predicted tumor volume, averaged over the three schedules; lower is better for all four. Every model has higher trajectory error in seed 2, which dominates that column's standard deviation. The dose-effect ratio compares predicted and simulated effects of sustained dosing at 0.6 versus 0.2; one is correct. $\lambda$ is the retention used to compute the exposure input; the simulator uses 0.78.}
\label{tab:p1}
\setlength{\tabcolsep}{4.5pt}
\begin{tabular}{lccccc}
\toprule
 & \multicolumn{3}{c}{Schedule-difference RMSE} & Trajectory & Dose-effect \\
\cmidrule(lr){2-4}
Model & Front-loaded & Break & Break, steps 5--8 & RMSE & ratio \\
\midrule
Zero-effect reference & 0.465 & 0.207 & 0.197 & --- & 0 \\
ADMIT, no exposure & $0.104\pm0.010$ & $0.081\pm0.009$ & $0.089\pm0.030$ & $0.62\pm0.16$ & $0.96\pm0.03$ \\
ADMIT, $\lambda=0.60$ & $0.096\pm0.003$ & $0.061\pm0.002$ & $0.064\pm0.014$ & $0.56\pm0.16$ & $0.89\pm0.03$ \\
ADMIT, $\lambda=0.78$ (true) & $0.102\pm0.012$ & $0.064\pm0.007$ & $0.064\pm0.010$ & $0.50\pm0.15$ & $0.84\pm0.03$ \\
ADMIT, $\lambda=0.90$ & $0.107\pm0.012$ & $0.064\pm0.006$ & $0.065\pm0.007$ & $0.48\pm0.12$ & $0.80\pm0.02$ \\
GRU, $\lambda=0.78$ & $0.123\pm0.020$ & $0.071\pm0.013$ & $0.066\pm0.013$ & $0.42\pm0.13$ & $0.85\pm0.04$ \\
\bottomrule
\end{tabular}
\end{table}

\textbf{ADMIT captures changes caused by dose timing.} With the correct clearance input, RMSE for the front-loaded and break differences is $0.102\pm0.012$ and $0.064\pm0.007$, respectively (Table~\ref{tab:p1}). These errors are 78\% and 69\% lower than the zero-effect reference. The example patient shows how this comparison unfolds: relative to steady dosing, tumor volume is lower while the break schedule gives more drug, the difference rises during dose reduction, and it falls after higher dosing resumes (Fig.~\ref{fig:p1}b,c). Tumor-volume predictions themselves still have error (trajectory RMSE 0.50). Errors shared across schedules can partly cancel when comparing their effects, so accurate treatment differences need not imply equally accurate individual trajectories.

\textbf{Clearance information helps during dose reduction, even when its rate is wrong.} Adding exposure lowers break-difference RMSE from $0.081\pm0.009$ to 0.061--0.064, a 21--25\% reduction seen in all three seeds for every tested retention value. Patient-bootstrap comparisons also support the improvement with the correct rate within each seed (Table~\ref{tab:p1-bootstrap}). The average gain is largest during steps 5--8, when error falls from 0.089 to 0.064--0.065. Front-loaded dosing shows no consistent benefit across seeds.

The two incorrect clearance rates perform similarly to the correct one, and their relative performance varies across seeds. A possible explanation is that the model already infers exposure from dose history: a linear readout of its history representation explains 96\% of the variation in simulator exposure without the added input and 97\% with it. The input may therefore make existing information easier to use. This interpretation applies to the tested setting: the wrong rate is shared by all patients, the exposure equation has the correct form, and the evaluated doses occur in training. Patient-specific clearance differences, an incorrect exposure equation and unfamiliar schedules remain untested.

\textbf{Better tumor-volume predictions can accompany weaker predicted treatment effects.} Every model correctly predicts that a higher sustained dose reduces tumor volume. ADMIT without exposure recovers nearly the full size of this benefit: the ratio of predicted to simulated effect is $0.96\pm0.03$, where one is correct. Adding exposure lowers the ratio to 0.89, 0.84 and 0.80 as assumed retention increases from 0.60 to 0.78 to 0.90. At the same time, tumor-volume RMSE improves from 0.62 without exposure to 0.56, 0.50 and 0.48 (Fig.~\ref{fig:p1} e,f). Both patterns hold in every seed. This tradeoff shows why outcome accuracy and treatment-effect accuracy must be assessed separately; the experiment does not establish its cause.

\textbf{A simpler model gives comparable average predictions.} The GRU's errors in the front-loaded and break differences are $0.123\pm0.020$ and $0.071\pm0.013$, compared with ADMIT's $0.102\pm0.012$ and $0.064\pm0.007$ using correct clearance. ADMIT has lower front-loaded error in two seeds, with no clear difference in the third; the break comparison changes direction across seeds. The GRU has lower tumor-volume error in every seed (0.42 versus 0.50 on average) and a similar dose-effect ratio (0.85 versus 0.84). It trains for five times as many epochs, and removing its exposure input further lowers its schedule-difference errors (Appendix~\ref{app:p1-ablations}). These results provide no clear accuracy advantage for diffusion on average predictions. Whether ADMIT's range of generated courses meaningfully represents uncertainty requires a separate calibration study.

The dosing experiment establishes that ADMIT can capture much of the response to changed treatment timing in this simulator, using only the available history. Its scope remains limited to three training seeds and one test cohort, and representation training also uses the simulator's noise-free target signal, which clinical records would not supply. It does not test adjustment for confounding, treatment plans that adapt to the patient's evolving condition, or clearance knowledge used to constrain generation. We examine this last use of physiological knowledge next.

\subsection{Applying an exposure limit during generation}
\label{sec:exp-projection}
Providing exposure as an input helps the model describe drug accumulation, but does not require its predictions to stay below a chosen limit. In a separate synthetic experiment, we apply such a limit directly. Before a generated state is used to continue the trajectory, a projection step adjusts the patient representation to reduce predicted exposure above a chosen limit of 3 (Eq.~\eqref{eq:projection}). Three adjustment iterations reduce violations across the generated states of 512 simulated patients over time. Under always treating, violations fall from 89.6\% to 1.0\% when the simulator's treatment-persistence parameter is 0.9, but from 92.8\% to 27.7\% when it is zero.

The constraint therefore reduces violations substantially, but enforcement remains incomplete and varies with the simulation setting. These single-run results measure compliance with a chosen exposure bound. Adjusting the generated state also changes the modeled treatment response, while the unchanged dosing plan may itself imply exposure above the limit. Reduced violations alone consequently establish neither correct treatment effects nor clinical safety. Rule enforcement must be assessed alongside intervention accuracy.

\subsection{Testing treatment responses with real clinical context}
\label{sec:exp-semisynthetic}
We next ask whether ADMIT can generate treatment responses when the surrounding patient measurements come from clinical records. Five streams of irregular observations from the eICU database \citep{pollard2018eicu} provide this context. We add two simulated outcomes and one shared treatment that can be on or off, so that outcomes under alternative plans remain known. This \emph{semi-synthetic} design includes 11,308 stays from 10,374 patients at 149 hospitals, split by patient into 7,909 training, 1,712 validation and 1,687 test stays.

The simulated outcomes depend on patient-specific patterns and 38 clinical context features. Treatment assignment depends on recent outcomes and clinical context; treatment then lowers the outcomes, with effects that fade over time and last for different durations in the two targets. The clinical context and untreated outcome paths remain fixed across treatment plans. Thus, the experiment evaluates a specified response mechanism embedded in real records; it does not estimate the effects of actual ICU treatments. The equations and training details are in Appendix~\ref{app:semisynthetic-settings}.

After eight hours of history, ADMIT generates outcomes for hours 9--24, updating the patient state one hour at a time. We compare never treating, always treating and an adaptive plan that starts treatment when predicted outcomes rise above an upper threshold and stops it when they fall below a lower threshold. Each decision uses the generated history available at that hour, and treatment given before forecasting continues to affect later outcomes.

ADMIT's treatment-effect RMSE is 0.488 for always treating and 0.357 for the adaptive plan, compared with 0.804 and 0.621 for predictions of no treatment effect (Table~\ref{tab:semisynthetic}). These results come from the latest completed configuration among 30 single-seed validation trials, evaluated on 1,712 validation stays. Each prediction uses one generated trajectory with fixed sampling noise. The results support further testing of treatment-response generation in clinical context, but do not yet establish prospective performance: the encoder uses the full observed record, some missing values are filled from later measurements, and imputation medians use the whole cohort. These sources of future or held-out information must be removed before a prospective evaluation.

\begin{table}[t]
\centering
\small
\caption{\textbf{Treatment comparisons with real clinical context and simulated outcomes.} Single training seed, 1,712 validation stays, 16 future hours and two standardized synthetic targets. Effects are relative to never treating after the forecast begins; the reference policy has zero effect by definition. The adaptive plan uses separate thresholds to start and stop treatment. Values describe the latest completed configuration and do not estimate variability across training runs.}
\label{tab:semisynthetic}
\begin{tabular}{lccc}
\toprule
Policy & Outcome RMSE & Effect RMSE & Effect correlation \\
\midrule
Never treat & 0.535 & --- & --- \\
Always treat & 0.601 & 0.488 & 0.679 \\
Adaptive & 0.498 & 0.357 & 0.724 \\
\bottomrule
\end{tabular}
\end{table}

\subsection{Examining treatment sensitivity in observed clinical outcomes}
\label{sec:exp-observational}
Finally, we turn to outcomes measured in clinical care: lactate, charted urine output and creatinine, with measurements of circulatory function (hemodynamics) as context. Here, only the outcomes under the treatments actually received are observed. We can therefore examine how predictions respond to a treatment query, but cannot check the alternative course against a known answer.

For 1,687 held-out eICU stays and one training seed, we generate 20 hours of outcomes after a nominal four-hour history. We compare no treatment with sustained norepinephrine-equivalent dosing of $0.5\,\mu$g/kg/min and crystalloid at $500$\,mL/h as illustrative model queries. Predicted differences between plans are small relative to error in predicting observed outcomes (Table~\ref{tab:clinical}). This scale comparison cannot establish that the treatments have little effect or rank the plans: alternative outcomes are unavailable, the two measures score different sets of entries, and the encoder again has access to the full observed window.

\begin{table}[t]
\centering
\small
\caption{\textbf{Sensitivity to treatment changes in observational clinical records.} One-seed normalized results for 1,687 eICU stays. Policy contrast is the mean absolute difference between sustained-treatment and no-treatment predictions. Factual RMSE scores observed entries; the final column is a descriptive scale ratio.}
\label{tab:clinical}
\begin{tabular}{lccc}
\toprule
Target & Policy contrast & Factual RMSE & Contrast/RMSE \\
\midrule
Lactate & 0.067 & 1.170 & 0.058 \\
Urine output & 0.120 & 0.466 & 0.256 \\
Creatinine & 0.018 & 0.398 & 0.045 \\
\bottomrule
\end{tabular}
\end{table}

Across these experiments, complementary observations improve supervised recovery of a hidden patient factor, dose-timing effects can be learned in a controlled simulator, and an explicit exposure limit reduces rule violations. The clinical-data experiments show how the same framework can be queried with more complex patient histories, while exposing the steps still needed for valid forecasting. The central lesson is that recovering patient information, predicting treatment effects and enforcing physiological rules are distinct requirements; progress on one must be checked against the others.

\section{Discussion}
Complementary observations improve supervised recovery of a simulated confounder, with a smaller improvement in treatment-contrast error. In a simulated chemotherapy setting with leak-free history, ADMIT predicts most of the effect of changing dose timing. An exposure input improves these predictions around a temporary dose reduction and tolerates a wrong clearance rate, but it shrinks the predicted size of a dose effect, and a deterministic recurrent model matches ADMIT on average predictions. Projection reduces rule violations but changes the transition law. These findings motivate separate measurements of representation recovery, intervention accuracy and constraint enforcement. Future work should assess generalization across patient populations and treatment settings, develop richer physiological constraints, and evaluate uncertainty calibration. Establishing clinical value will require further validation using only information available at each treatment decision.

\begingroup
\small
\urlstyle{same}
\raggedright
\setlength{\bibsep}{3pt plus 1pt}
\bibliography{reference}
\endgroup

\clearpage
\appendix
\section{Dosing-schedule experiment: settings, checks and ablations}
\label{app:p1}
This appendix lists the settings and validity checks of the dosing-schedule experiment (Section~\ref{sec:p1}) and ablations that support, but do not change, its conclusions.

\subsection{Settings and checks}
Table~\ref{tab:p1-settings} lists the settings. All conditions share one Stage-1 checkpoint per training seed, and their transition settings differ only in the exposure input. Training uses factual trajectories only.

\begin{table}[h]
\centering
\small
\caption{\textbf{Dosing-schedule experiment settings.}}
\label{tab:p1-settings}
\begin{tabular}{p{0.16\linewidth}p{0.78\linewidth}}
\toprule
Simulator & Tumor model of Section~\ref{sec:experiments} with one continuous drug and the hidden comorbidity off; on the logit scale, each dose is centered at 0.9 times the previous dose with noise standard deviation 0.8; retention $\lambda=e^{-1/4}$; 12 history and 12 forecast steps; block size four. \\
Patients & 1,000 training, 200 validation and 300 test patients, fixed for all conditions and training seeds. \\
Stage~1 & Encoder of Section~\ref{sec:method} with time attention and a local kernel; 16-dimensional latent state; 60 epochs, batch size 128, learning rate $10^{-3}$; no exposure head; also supervised with the simulator's noise-free target signal. \\
Transition & 40 epochs, batch size 128, learning rate $10^{-3}$, hidden size 64, 100 diffusion steps; rollout-loss weight three over four free-running steps; no condition dropout; guidance scale one. \\
GRU baseline & Deterministic recurrent network with hidden size 128 whose input is the latent state, dose and exposure; 200 epochs; trained on true previous states and on its own previous predictions. \\
Evaluation & 64 ADMIT samples with sampling noise shared across schedules; simulator references average 32 runs with process noise shared across schedules. \\
\bottomrule
\end{tabular}
\end{table}

Table~\ref{tab:p1-checks} reports three checks per training seed. The first tests the leak-free encoding. Every observation after the last history step is replaced with another patient's values plus noise, and we record the largest change in the encoded history. The leak-free history does not change, whereas the full-record encoder used in the other experiments changes substantially; removing the leak required both marking later observations as missing and setting them to zero. The second check confirms that the model's exposure with $\lambda=0.78$ matches the simulator's drug level on factual doses and on every planned schedule. The third is the linear exposure readout discussed in Section~\ref{sec:p1}. Each schedule dose level (0.2--0.6) makes up 9.5--9.9\% of factual training doses (within $\pm0.05$).

\begin{table}[h]
\centering
\small
\caption{\textbf{Validity checks per training seed.} History change: largest absolute change in the encoded history after all observations following the last history step are replaced. Exposure error: largest absolute difference between the model's exposure ($\lambda=0.78$) and the simulator's drug level over factual doses and all planned schedules. Probe $R^2$: test-set fraction of variance in the simulator's exposure explained by a linear regression on the history summary $R_t$, for each exposure input.}
\label{tab:p1-checks}
\begin{tabular}{cccccccc}
\toprule
 & \multicolumn{2}{c}{History change} & & \multicolumn{4}{c}{Probe $R^2$} \\
\cmidrule(lr){2-3}\cmidrule(lr){5-8}
Seed & Leak-free & Full record (max / mean) & Exposure error & None & $\lambda=0.60$ & $\lambda=0.78$ & $\lambda=0.90$ \\
\midrule
0 & 0 & 6.75 / 0.51 & $<3\times10^{-7}$ & 0.962 & 0.969 & 0.971 & 0.973 \\
1 & 0 & 4.98 / 0.54 & $<3\times10^{-7}$ & 0.961 & 0.971 & 0.969 & 0.971 \\
2 & 0 & 5.09 / 0.48 & $<3\times10^{-7}$ & 0.961 & 0.969 & 0.973 & 0.969 \\
\bottomrule
\end{tabular}
\end{table}

The four-step joint blocks do not introduce the treatment-selection issue described in Eq.~\eqref{eq:blockselection}: doses do not depend on tumor states, so the middle factor does not vary with $l_1$. This experiment tests response to dose timing without testing adjustment for confounding. In the training data, 9\% of four-step dose blocks fall and 6\% rise by at least 0.3. These checks document coverage of the evaluated dose levels and changes.

Tumor volume $V$ is read from the generated target vital, which the simulator sets to $2+0.15V$. The no-exposure condition sets the exposure input to zero. All ADMIT conditions share the encoder and decoder trained once per seed; the GRU uses the same decoder and generates one next state per step. Schedule-difference RMSE pools patients and forecast steps, comparing each alternative schedule's tumor volume minus the steady schedule's volume with the corresponding simulator difference. Trajectory RMSE averages tumor-volume RMSE over the three schedules. The dose-effect ratio divides the average predicted difference between sustained doses of 0.6 and 0.2 by the simulated difference.

\subsection{Ablations}
\label{app:p1-ablations}
\textbf{Paired bootstrap.} Table~\ref{tab:p1-bootstrap} compares pairs of models within each training seed. The intervals are narrow compared with the differences between seeds, so the training seed, not the choice of test patients, drives most comparisons in Section~\ref{sec:p1}.

\begin{table}[h]
\centering
\small
\caption{\textbf{Paired patient-bootstrap differences in schedule-difference RMSE.} For each training seed, the 300 test patients are resampled with replacement 4,000 times, with the same resample for both models. Entries give the first model's RMSE minus the second's, with a 95\% percentile interval in brackets; negative values favor the first model. ADMIT and GRU rows without a stated input use the true $\lambda=0.78$.}
\label{tab:p1-bootstrap}
\footnotesize
\setlength{\tabcolsep}{3pt}
\begin{tabular}{llccc}
\toprule
Comparison & Difference & Seed 0 & Seed 1 & Seed 2 \\
\midrule
No exposure $-$ true $\lambda$ & Front-loaded & $-$0.019 [$-$0.023, $-$0.014] & 0.012 [0.006, 0.017] & 0.014 [0.009, 0.018] \\
 & Break & 0.003 [0.001, 0.005] & 0.016 [0.013, 0.018] & 0.033 [0.029, 0.036] \\
$\lambda=0.60$ $-$ true $\lambda$ & Front-loaded & $-$0.018 [$-$0.021, $-$0.016] & 0.003 [$-$0.000, 0.005] & $-$0.002 [$-$0.005, 0.001] \\
 & Break & $-$0.012 [$-$0.014, $-$0.011] & 0.002 [0.001, 0.003] & 0.001 [$-$0.001, 0.003] \\
$\lambda=0.90$ $-$ true $\lambda$ & Front-loaded & $-$0.004 [$-$0.006, $-$0.001] & 0.005 [0.003, 0.007] & 0.016 [0.013, 0.019] \\
 & Break & $-$0.012 [$-$0.014, $-$0.011] & $-$0.001 [$-$0.003, 0.000] & 0.012 [0.010, 0.013] \\
GRU $-$ ADMIT & Front-loaded & 0.004 [$-$0.002, 0.010] & 0.016 [0.012, 0.021] & 0.045 [0.038, 0.054] \\
 & Break & $-$0.005 [$-$0.008, $-$0.002] & $-$0.001 [$-$0.004, 0.002] & 0.027 [0.024, 0.030] \\
GRU, no exposure $-$ GRU & Front-loaded & $-$0.010 [$-$0.014, $-$0.007] & $-$0.005 [$-$0.009, $-$0.001] & $-$0.016 [$-$0.021, $-$0.011] \\
 & Break & $-$0.007 [$-$0.009, $-$0.004] & $-$0.003 [$-$0.005, $-$0.000] & $-$0.021 [$-$0.024, $-$0.018] \\
\bottomrule
\end{tabular}
\end{table}

\textbf{GRU without exposure.} Removing the exposure input lowered the GRU's schedule-difference errors in every seed (last rows of Table~\ref{tab:p1-bootstrap}): break RMSE is $0.061\pm0.004$ without exposure against $0.071\pm0.013$ with it, and front-loaded RMSE $0.113\pm0.015$ against $0.123\pm0.020$. Without exposure, its trajectory RMSE is $0.44\pm0.09$ and its dose-effect ratio $0.89\pm0.03$. Exposure therefore helped ADMIT's break predictions but not the GRU's.

\textbf{Shared sampling noise.} An earlier evaluation drew independent sampling noise for each schedule. The Monte Carlo error of ADMIT's break difference, the part due to using a finite number of samples, was then 0.119 against a total RMSE of 0.139, which hid the differences between conditions. Sharing sampling noise across schedules, as the simulator shares process noise, reduced this error to 0.005 with 64 samples. All reported results use shared noise.

\textbf{Training length.} On seed 0, training the transition for 120 instead of 40 epochs raised front-loaded RMSE for all four ADMIT conditions (0.123--0.129 against 0.095--0.114) and raised break RMSE for three of the four. This check was scored on the test patients; the 40-epoch setting was kept.

\section{Additional experimental settings}
\label{app:additional-experiments}
\subsection{Hidden-confounder recovery}
The first experiment in Section~\ref{sec:experiments} adds a hidden autoregressive comorbidity $H$ to the tumor simulator. Five irregular modalities provide observations over 40 steps with 10\% missingness. Each of three seeds generates 1,024 training and 256 validation subjects; the archived sweep evaluates the validation cohort. Both training stages run for 15 epochs, with block size five and rollout-loss weight three. Both models train a comorbidity readout using simulator labels; context reconstruction is enabled only in the multimodal arm.

\subsection{Semi-synthetic outcomes and treatment assignment}
\label{app:semisynthetic-settings}
For Section~\ref{sec:exp-semisynthetic}, untreated outcomes $U_{j,t}$ combine a spline, a patient-specific Gaussian process and a random function of 38 context features. Assignment probability is $p_t=\operatorname{sigmoid}(-1+2\bar Y^{\rm raw}_t+2f_A(X_t))$, where $\bar Y^{\rm raw}_t$ averages both outcomes over the current and up to three preceding hours, and $f_A$ is a fixed random-feature function. Treatment adds
\begin{equation}
 Y^{\rm raw}_{j,t}=U_{j,t}+\sum_{s=\max(1,t-w_j)}^{t-1}
 \frac{\beta_j A_s p_s}{\sqrt{t-s}},\qquad
 (\beta_1,\beta_2)=(-1,-0.75),\quad(w_1,w_2)=(20,7),
 \label{eq:semisynthetic}
\end{equation}
where $p_s$ is the simulator assignment probability, recomputed using each intervention arm's history. Outcomes are standardized using factual training statistics. Real context and untreated paths remain fixed across arms.

The representation and one-step transition train for 30 and 100 epochs, respectively, with rollout-loss weight three and no propensity weighting or exposure guard. Eight history hours precede forecasts for hours 9--24. The adaptive policy starts off, turning on above 0.208 and off below $-0.057$ in mean standardized outcomes (training 60th/40th percentiles). Decisions use prefix-decoded outcomes; final grid decoding can revise earlier outputs using the generated suffix. Earlier treatments retain carryover after the forecast begins. Results use one fixed-noise rollout and report the latest completed configuration among 30 single-seed validation trials. Full-record encoding, backward filling and cohort-wide imputation medians require correction before prospective evaluation.

\section{Rollout and exposure specification}
\label{app:rollout-exposure}
Rollout starts from the history prefix and preceding treatments. Each generated block is optionally constrained, appended to the sequence and used to update the history summary. Actions follow a fixed plan or a policy evaluated on generated history, preserving $A_t\rightarrow Z_{t+1}$ alignment. All reported runs use guidance scale one; amplified guidance would alter the fitted distribution.

Where a specified exposure model applies, carryover follows $C_{t+1}=\lambda C_t+A_t$, with $\lambda=\exp(-\Delta/\tau)$ for step length $\Delta$ and time constant $\tau$. The tumor simulator uses $\tau=4$ simulation steps, so each step keeps $\lambda=e^{-1/4}\approx0.78$ of the exposure; Section~\ref{sec:p1} also supplies wrong values of $\lambda$ to test a misspecified rate. This feature encodes an assumed mechanism; it neither establishes pharmacological validity nor supplies treatment support. The semi-synthetic run omits it.

\end{document}